\documentclass[12pt]{article}
\usepackage{fullpage}
\usepackage{graphicx}
\usepackage{float}
\usepackage{amsmath}
\usepackage{amssymb}
\usepackage{amsthm}
\usepackage{natbib}
\usepackage{url}
\usepackage{threeparttable}
\usepackage[hmargin=1.0in,vmargin=1.0in]{geometry}
\usepackage{authblk}
\usepackage{pdfpages}
\usepackage{color}
\usepackage{hyperref}

\usepackage[markup=underlined]{changes}
\usepackage{todonotes}

\definechangesauthor[color=blue]{YB}
\definechangesauthor[color=red]{GM}
\usepackage{fancyhdr}
\usepackage{latexsym}
\usepackage{verbatim}
\usepackage{multirow}
\usepackage{caption} 
\usepackage{natbib}
\usepackage{listings}
\usepackage[export]{adjustbox}
\usepackage{titlesec}
\titleformat*{\section}{\LARGE\bfseries}
\titleformat*{\subsection}{\Large\bfseries}
\titleformat*{\subsubsection}{\large\bfseries}

\usepackage{setspace}

\begin{document}
%\listofchanges[style=<list|summary>]
\title{Depth video data-enabled predictions of longitudinal dairy cow body weight using thresholding and Mask R-CNN algorithms}

\author[1]{Ye Bi}
\author[1]{Leticia M.Campos}
\author[2]{Jin Wang}
\author[2]{Haipeng Yu}
\author[1]{Mark D.Hanigan}
\author[1, 3, *]{Gota Morota}
\affil[1]{School of Animal Sciences, Virginia Tech, Blacksburg, VA, 24061 USA}
\affil[2]{Department of Animal Sciences, University of Florida, Gainesville, FL, 32611 USA}
\affil[3]{Center for Advanced Innovation in Agriculture, Virginia Tech, Blacksburg, VA, 24061 USA}

\date{}

\maketitle

\noindent 
$^{*}$ Corresponding author \\ 

\noindent
Running title: Deep learning prediction of dairy cattle body weight \\


\noindent
ORCID: 0000-0001-7871-5856 (YB), 0000-0003-3726-5670 (LMC), 0009-0001-4389-4396 (JW), 0000-0002-8923-9733 (HY), 0000-0003-5639-9677 (MDH), and 0000-0002-3567-6911 (GM) \\
 
\noindent
Email addresses: yebi@vt.edu (YB), leticiamc@vt.edu (LMC), jin.wang@ufl.edu (JW), haipengyu@ufl.edu (HY), mhanigan@vt.edu (MDH), and morota@vt.edu (GM)

\newpage
\doublespacing

\section{Abstract}
Monitoring cow body weight is crucial to support farm management decisions due to its direct relationship with the growth, nutritional status, and health of dairy cows. Cow body weight is a repeated trait, however, the majority of previous body weight prediction research only used data collected at a single point in time. Furthermore, the utility of deep learning-based segmentation for body weight prediction using videos remains unanswered. Therefore, the objectives of this study were to predict cow body weight from repeatedly measured video data, to compare the performance of the thresholding and Mask R-CNN deep learning approaches, to evaluate the predictive ability of body weight regression models, and to promote open science in the animal science community by releasing the source code for video-based body weight prediction. An Intel RealSense D435 camera was installed on Virginia Tech dairy complex to collect top-view videos of 10 lactating Holstein cows twice a day for 28 days after milking sessions and 2 non-lactating, pregnant Jersey cows once a week upon calving. At the same time, the ground truth body weight records were collected using a walk-over weighing system. A total of 40,405 depth images and depth map files were obtained. Three approaches were investigated to segment the cow's body from the background, including single thresholding, adaptive thresholding, and Mask R-CNN. Four image-derived biometric features, such as dorsal length, abdominal width, height, and volume, were estimated from the segmented images and fitted using ordinary least squares, ridge regression, least absolute shrinkage and selection operator, and linear mixed models. Two cross-validation designs, forecasting and leave-three-cows-out, were used to evaluate prediction performance. The Pearson correlation between image-derived biometric features and scale-based body weight ranged from 0.74 to 0.95. On average, the Mask-RCNN approach combined with a linear mixed model resulted in the best prediction coefficient of determination and mean absolute percentage error of 0.98 and 2.03\%, respectively, in the forecasting cross-validation. The Mask-RCNN approach was also the best in the leave-three-cows-out cross-validation, followed by adaptive and single thresholding. The prediction coefficients of determination and mean absolute percentage error of the Mask-RCNN coupled with the linear mixed model were 0.90 and 4.70\%, respectively. Our results suggest that deep learning-based segmentation improves the prediction performance of cow body weight from longitudinal depth video data.

\newpage 
\section{Background}
The body weight of dairy cattle varies on a daily basis. Regular monitoring of body weight is important because estimates of body weight can provide insight into management factors that may affect milk production, feed efficiency, energy balance, growth, and health status. For example, a negative energy balance can potentially indicate an inadequate diet or health and reproductive problems. For this reason, individual monitoring of body weight throughout a cow's lifespan is essential to make better management decisions. However, body weight monitoring is not performed as often as it should be because it is labor- and time-consuming when measured manually or cost-prohibitive if using an automated walk-over weighing system. In addition, the trend toward larger average farm sizes provides more opportunities for increasing the efficiency of identification of cow health problems. The use of a low-cost, non-contact, stress-free 3 dimensional (3D) depth sensor camera coupled with computer vision analysis has been proposed as an alternative method to address the challenges associated with assessing cow morphometric traits, including predicting body weight \citep{norton2017developing}. Computer vision is considered one of the sub-fields in a broader effort of precision livestock farming, which aims to integrate advances in engineering and technology, and monitor animals at individual levels to positively intervene in farm management \citep{morota2018big}. Although earlier computer vision applications for dairy cows involve the simultaneous use of multiple 2D cameras, a detailed calibration process is required to synchronize them \citep{tasdemir2011determination,le2019volume}. A 3D depth sensor camera recently used in computer vision provides distances between the camera and objects in addition to standard two-dimensional information, including body contours. Height- and volume-derived 3D images have been shown to improve the performance of body weight predictions as assessed by model goodness of fit \citep{kuzuhara2015preliminarily,hansen2018automated, martins2020estimating,xavier2022use,le2022three} and cross-validation \citep{nir20183d, song2018automated,le2019volume}. 

Body weight is a repeated trait. However, previous body weight prediction research has used data collected at a single point in time without repeated measurements from the same cow.  This type of experimental design does not allow evaluation of the predictive performance of models for observations that vary over time. Previous studies in dairy cattle that collected repeated body weight measurements used limited sample sizes, limited time points, manually selected the best image from each video, or did not employ cross-validation to evaluate predictive ability \citep{martins2020estimating,xavier2022use,le2022three}. Therefore, an alternative approach is needed that can better explore longitudinal records and improve body weight prediction.

Deep learning has been recognized as mainstream in image and video data analysis. Yet, the usefulness of deep learning for dairy cow body weight prediction has not been assessed. Image segmentation is often a critical first step in computer vision to separate the cow from background information. The quality of cow biometric features extracted from segmented images heavily influences the predictability of traits of interest. Among various deep learning models, Mask R-CNN is used for instance segmentation, object detection, and object classification \citep{he2017mask}. Although Mask R-CNN has been witnessing growth in cow detection \citep{qiao2019cattle,xu2020automated, salau2020instance,bello2021enhanced,xiao2022cow}, research evaluating the utility of Mask R-CNN for body weight prediction is in its early stages and has not been evaluated in great detail. Additionally, applying deep learning-based computer vision to animal video data often involves extensive data analysis using custom computer codes. Open science promotes research products and processes that are accessible to all levels of the community \citep{nosek2015promoting}. One way to practice open science is to make the source code used in research publications available to the public. However, it has been pointed out that the adoption of open source code in animal science is lagging behind other disciplines \citep{munoz2022seven}. In particular, none of the computer vision-based body weight prediction work in the published literature to date has released its source code. The lack of available open source code makes animal computer vision data analysis processes less transparent and less reliable. Breaking down these barriers requires making open source code used in analysis accessible to the animal science community without any restrictions. 

Therefore, the objectives of this study were 1) to predict cow body weight from repeatedly measured video data, 2) to compare the performance of automated thresholding and Mask R-CNN approaches, 3) to evaluate the predictive ability of body weight regression models, and 4) to release open source code for depth video-based body weight prediction to the animal science community.

\newpage 
\section{Materials and methods}
\subsection{Animal experiments}
Animal handling and media recording were approved and conducted in accordance with the Virginia Tech Institutional Animal Care and Use Committee. A total of 12 multiparous animals were used. Ten lactating Holstein cows (286 days in milk) with average initial body weight (SD) of 793 (148) kg and two pregnant non-lactating Jersey cows with initial body weight of 552 (33) kg at 12- and 21-days ante-parturition from Kentland Farm Dairy Complex (Virginia Tech, Blacksburg, VA) were used in the current study. Cows were housed in a free-stall barn, milked twice daily, fed in \textit{ad libitum} regime with a common diet once daily, and had free access to water. Data were collected in 2020. Lactating animal body weight were recorded daily for 28 days after cows left the milking parlor following the 12:00 am and 12:00 pm milking sessions. Non-lactating animal body weight was recorded once a week. Both animals were handled and restrained individually over the walk-through scale, where the scale-based body weight was concomitantly and automatically recorded.

\subsection{Video acquisition setup}
An Intel RealSense D435 depth sensor camera (Intel, Santa Clara, CA, USA) connected to a laptop via a USB 3.1 cable was used to obtain depth video data from cows in this study. The sensor camera provides 87$^\circ$ horizontal and 58$^\circ$ vertical fields of view and uses two stereo image sensors to determine depth. A single-sensor camera was placed in a temperature-controlled box to maintain camera temperatures within the normal operating range. The box was mounted 2.95 meters above a one-way exit lane equipped with a walk-over weighing system between the milking parlor and pen housing, positioned straight down to capture top views of the cows. Equipment installation and location did not interfere with daily farm activities. Each cow was manually restrained to collect depth data in a short video format of 10-12 s at 30 frames per second and a resolution of 848$\times$480. Although the cow's body motion was stable throughout the videos, the head and tail were relatively active. The total size of all the video data was 50.1 GB. During the video recording, the cow's body weight was automatically recorded by a walk-through weighing system (Afimilk, Kibbutz Afikim, Israel). Lactating animals were identified via pedometers located at the animals' ankle. Due to pedometer absence in non-lactating animals body weight, identification, and time recordings using the same scale were executed after milking time and manually recorded.

\subsection{Image segmentation}
After the completion of on-farm video acquisition, off-farm data process was initiated. The rs-convert program was used to convert .bag video files into PNG depth images and comma-separated value (CSV) depth map files \citep{realsenseGitHub}. The size of the video data increased to 103.2 GB after converting to PNG and CSV. Different colors in each depth image are proportional to distances from the camera to the object. The corresponding CSV file of the same size contains the distance in meters for each pixel. Three image segmentation techniques, single-thresholding, adaptive-thresholding, and Mask R-CNN, were used to segment the cow from its background. The background was considered extraneous information in the images that does not represent the cows' body (i.e., constraint fences).

\subsubsection{Single-thresholding}
\begin{figure}[H]
    %\hspace*{-4cm} 
    \centering
    \includegraphics[scale=0.6]{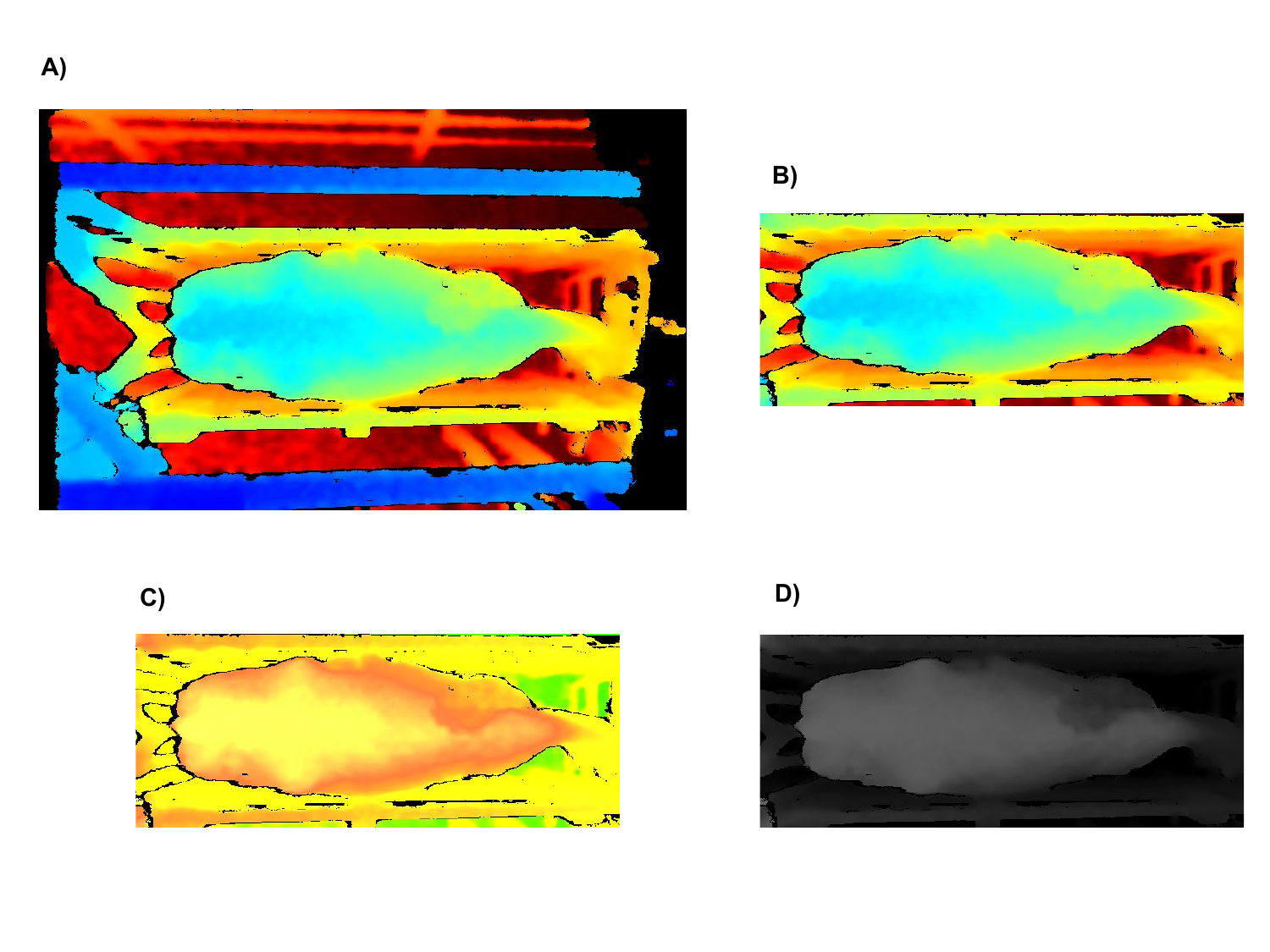}
    \caption{Image conversion process. A) The original depth image. B) Cropped depth image. C) The RGB color model was converted to the HSV color model. D) Hue was extracted and used for subsequent analyses. }
\label{HSV}
\end{figure}

\begin{figure}[H]
    %\hspace*{-4cm} 
    \centering
    \includegraphics[scale=0.60]{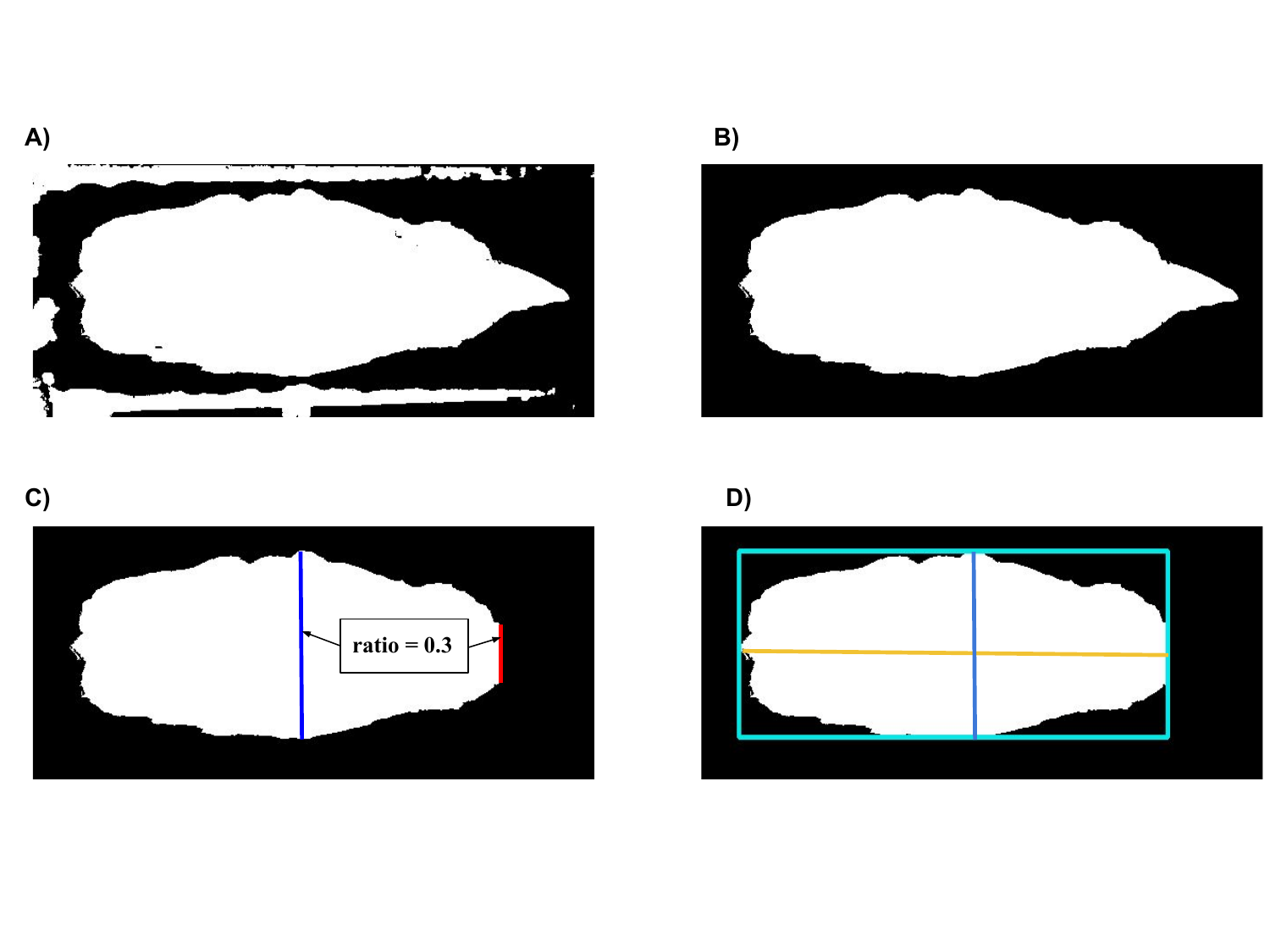}
    \caption{Neck removal process. A) Binary image after thresholding. B) The initial segmentation that includes the neck. C) The ratio of body width was used to remove the neck. Blue line represents the column location that returns the maximum number of white pixels across all columns. Red line represents the column location that provides the ratio of 0.3 relative to the blue line. D) The final cow body area with the bounding box. The cyan box, yellow line, and blue line denote the bounding box,  dorsal length, and abdomen width, respectively.}
\label{heads}
\end{figure}

The OpenCV library (version 4.7.0) was used in Python (version 3.10.9) for both thresholding methods. First, all depth images were cropped while maintaining the same ratio to remove the unnecessary surrounding area (i.e, background) while keeping the body part of the cow intact. The cropped images were converted from RGB (red, green, and blue) to HSV (hue, saturation, and value) color models (Figure \ref{HSV}). The hue channel was extracted to segment the cow's body \citep{kadlec2022automated}. Hue values below the threshold were converted to black and those above were converted to white. Single-thresholding is often used in the literature for image-based body weight prediction \citep{martins2020estimating, yu2021forecasting}. In single-thresholding, the mean of all hue values was applied to all images as a fixed threshold. After thresholding, the OpenCV findcontour function was used to identify and isolate the largest area connecting all contiguous points of equal value, commonly known as the contour. The contour was then drawn on a black mask to remove depth-blank areas (i.e., missing data chunks) in the segmented cow binary image. The body width ratio was used to remove the neck in each image (Figure \ref{heads}). Specifically, the number of white pixels in each column was calculated and the maximum number of white pixels across all columns was kept. We then calculated the ratio of the total number of white pixels in each column to the maximum value, starting from the center to the head side of the image. If the ratio was less than 0.3 (according to a preliminary analysis), we added several columns of black pixels to remove neck and also ensured that the image size remained the same. Finally, the contour was identified and the bounding box was drawn on the binary image. The bounding box and depth information associated with the identified contour were retained for body weight regression.

\subsubsection{Adaptive-thresholding}
A potential drawback of the single-thresholding approach is that the best threshold can vary dramatically from image to image due to differences in cow position and lighting conditions. For example, an inadequate threshold will result in incomplete background removal, often resulting in the cow's body being connected to the fence rail, while over-removal of the background will reduce the area of the cow's body part. Lighting conditions at noon and midnight also affect the pixel intensity of an image. To address this challenge, single-thresholding was extended by exploring the optimal hue threshold from minimum to maximum in the adaptive-thresholding approach rather then keeping a fixed threshold. We started with the lowest hue threshold for each image to create a binary image. We drew the bounding box using the OpenCV boundingRect function around the largest contour to determine an appropriate threshold. If the coordinates of four corners were less than or equal to five pixels from the edge, the hue threshold was increased by one unit until the four corners were greater than five pixels from the edge of image, because in this case the entire cow body contour is obtained without being connected to the fence rails. The OpenCV findcontour function was then used to find the contour with the largest area, and the contour was drawn on a black mask to remove empty areas in the segmented cow binary image. Finally, we used the OpenCV minAreaRect function to draw the minimum bounding rectangle of the cow's body contour. The same image processing step was used to remove the neck in each image, identify the contour, and draw the final bounding box (Figure \ref{heads}).

\subsubsection{Mask R-CNN}
\begin{figure}[H]
    %\hspace*{-4cm} 
    \centering
    \includegraphics[scale=0.90]{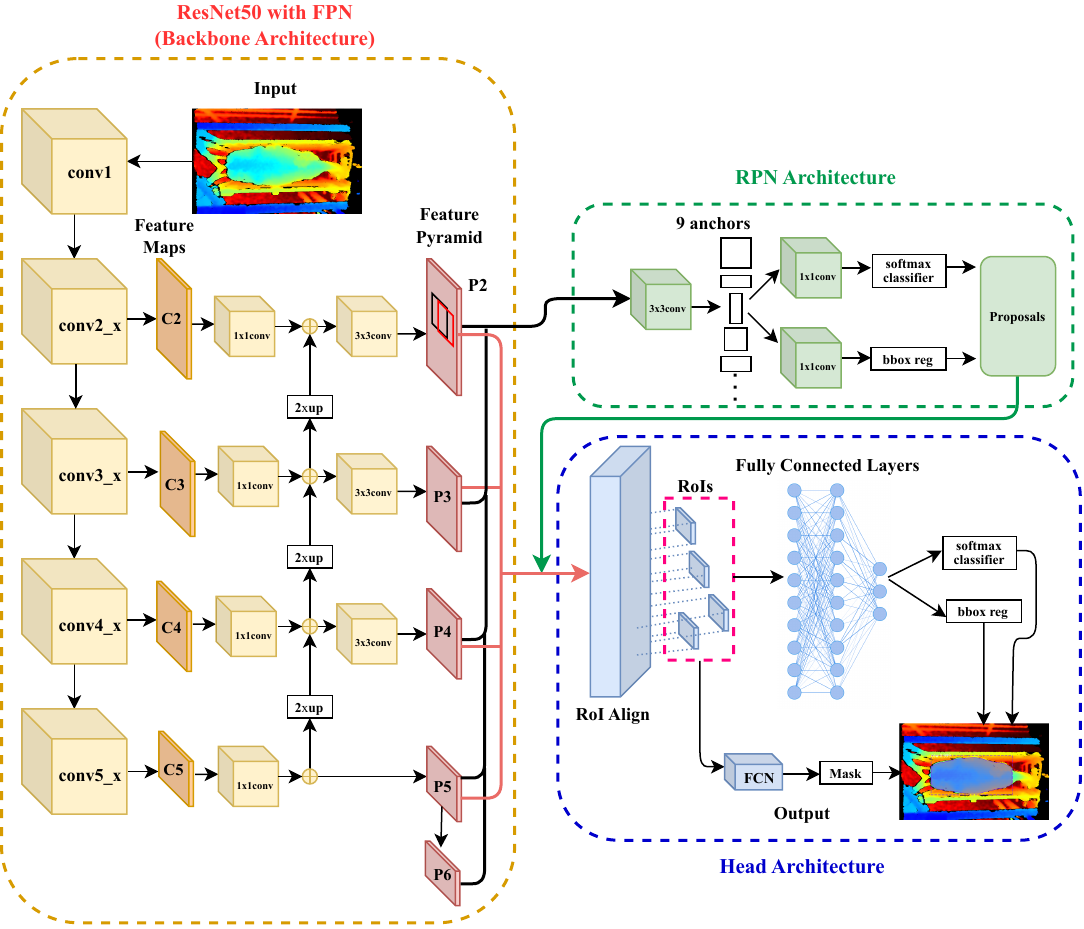}
    \caption{Mask R-CNN architecture used in the current study. conv1: initial convolutional block. conv\_x: multiple residual convolutional blocks. C2-C5: feature maps generated from each residual block. P1-P5: feature maps generated after the feature pyramid network (FPN). RPN: regional proposal network. RoI: region of interest. FCN: fully connected neural network. Bbox reg: bounding box regression.}
\label{mrcnn_arch}
\end{figure}

Mask R-CNN is developed on top of Faster R-CNN \citep{ren2015faster} by adding a new branch that generates a binary mask on each region of interest (RoI). In addition, Mask R-CNN also incorporates RoIAlign, which enhances feature alignment and aids in accurate mask generation. The overall architecture of Mask R-CNN is shown in Figure \ref{mrcnn_arch}. It includes the backbone network to extract features from the input image, followed by a region proposal network (RPN) that generates object proposals, and a head network that produces the final outputs of the model, including object class labels, bounding box regression, and binary mask predictions for object instances in the input image. 

\noindent
\textbf{Backbone}: All depth images were provided to the backbone of Mask R-CNN after being resized to 1024 $\times$ 1024 $\times$ 3 with zero padding. The backbone of Mask R-CNN was implemented using ResNet50, a 50-layer residual convolutional neural network model, to perform feature extraction over an entire image \citep{he2016deep}. Its architecture includes five stages that extract features from low level (edges and corners) to high level (rich semantics). The first stage includes a convolutional layer and a max-pooling layer. The second through fifth stages contain three, four, six, and three iterations of three-layer convolutional blocks with different designs, respectively, followed by an average pooling layer. In total, there are 48 convolutional layers, one max pooling layer, and one average pooling layer. Stages 2 to 4 in ResNet50 include skip connections, which allow information to be passed directly from one stage to another without going through the intermediate layers, to avoid a vanishing gradient problem, improve prediction, and facilitate convergence. The bottom-up pathway in ResNet50, coupled with the rectified linear unit activation function from the last residual block of each stage, produces a pyramid of feature maps. They were further processed via the top-down pathway to develop the feature pyramid network (FPN) with the aim of improving segmentation performance \citep{lin2017feature}. \\

\noindent
\textbf{Region proposal network}: The feature maps generated from the backbone and FPN are fed into a region proposal network (RPN), which is a fully convolutional network, to generate the region proposals \citep{ren2015faster}. It uses sliding windows over the feature maps to detect whether the target object (i.e., cow) is present within the region proposals to define ROIs and determine their coordinates. In particular, anchors (rectangles) with nine different sizes, three scales, and three aspect ratios were used. In total, about 21,000 anchors were used per image, since the feature map size after feature extraction was 40 $\times$ 60. The softmax function was used for each anchor to classify the cow as foreground or background. Here, foreground means that the cow is located in that particular anchor. For anchors with the most possibilities to contain the cow, the bounding box size and location were corrected to get a better fit. Non-maximum suppression was applied to remove bounding boxes that overlapped with other bounding boxes with higher foreground scores. The final set of RoIs are regions in the feature map that contain the cow. \\

\noindent
\textbf{Region of interest align}: Region of interest align (RoIAlign) was used to properly project the RoIs onto the extracted features to produce better pixel segmentation results \citep{he2017mask}. It adjusts the size of the bounding box to a fixed size of 7 $\times$ 7 and splits the proposal into 7 $\times$ 7 sub-regions within each bin. The bilinear interpolated value of four top points from the image grid was used to keep a single value per bin. \\

\noindent
\textbf{Mask head}: After the RoIAlign step, the fixed size RoIs were fed into a fully connected layer for object classification and bounding box regression. The RoIs were also fed to a small fully convolutional network for cattle body segmentation via mask prediction. Specifically, at each pixel location in the RoI, the fully convolutional neural network produced a mask prediction indicating the probability of that pixel belonging to the object class. The mask predictions are then post-processed to obtain the final cow segmentation by thresholding the mask probabilities and assigning each connected component to a unique object instance (i.e., cow). This process is equivalent to generating a mask for each RoI, allowing for pixel-level image segmentation. \\

\noindent
\textbf{Model training}: A multi-task loss function was used to determine whether the training process of bounding box regression, classification, and mask prediction in a Mask R-CNN model was complete. It was calculated as $L = L_{c} + L_{b} + L_{m}$, where $L_{c}$ is the classification loss, $L_{b}$ is the bounding box regression loss, and $L_{m}$ is the mask loss. The training process of Mask R-CNN was considered complete when the loss converged.  The Mask R-CNN model was pre-trained on the COCO dataset, which is widely used for object detection, image segmentation, and image annotation \citep{cocodataset}. It contains more than 300,000 annotated images. In addition to the COCO dataset, we randomly selected 300 depth images per cow per time point per day to include more specific situations for cow identification. The use of a pre-trained model is known as transfer learning, where we transferred the parameters from the previous training as initial values. We used Labelme to manually label the ground truth for the cow contour from the background  \citep{wada2018labelme}. These labeled cow images were added to the training dataset to fit the Mask R-CNN model. Segmentation performance was evaluated using a confidence score, which ranges from 0 to 1, with higher values indicating better segmentation accuracy \cite{matterport_maskrcnn_2017}. The Mask R-CNN was implemented using Keras \cite{matterport_maskrcnn_2017}. We used a batch size of 1, a step number of 300, an epoch of 42, learning rate 0.001. Since the original version was based on Python 3.4, TensorFlow 1.3, and Keras 2.0.8, we manually upgraded to Python 3.9, TensorFlow 2.9, and Keras 2.9, and updated the Mask R-CNN source code.

\noindent
\textbf{Prediction}
The estimated mask R-CNN parameters were used to segment cow contours in the remaining depth images. To reduce computational time, we chose to predict a subset of depth images for contour segmentation. Specifically, we excluded the first two frames in each video and retained one frame every 15 frames, resulting in predictions of 2 frames per second. The total number of newly predicted frames was (40405 - 300)/15 = 2674. 

\subsection{Biometric features}
\begin{figure}[H]
    %\hspace*{-4cm} 
    \centering  
    \includegraphics[scale=0.70]{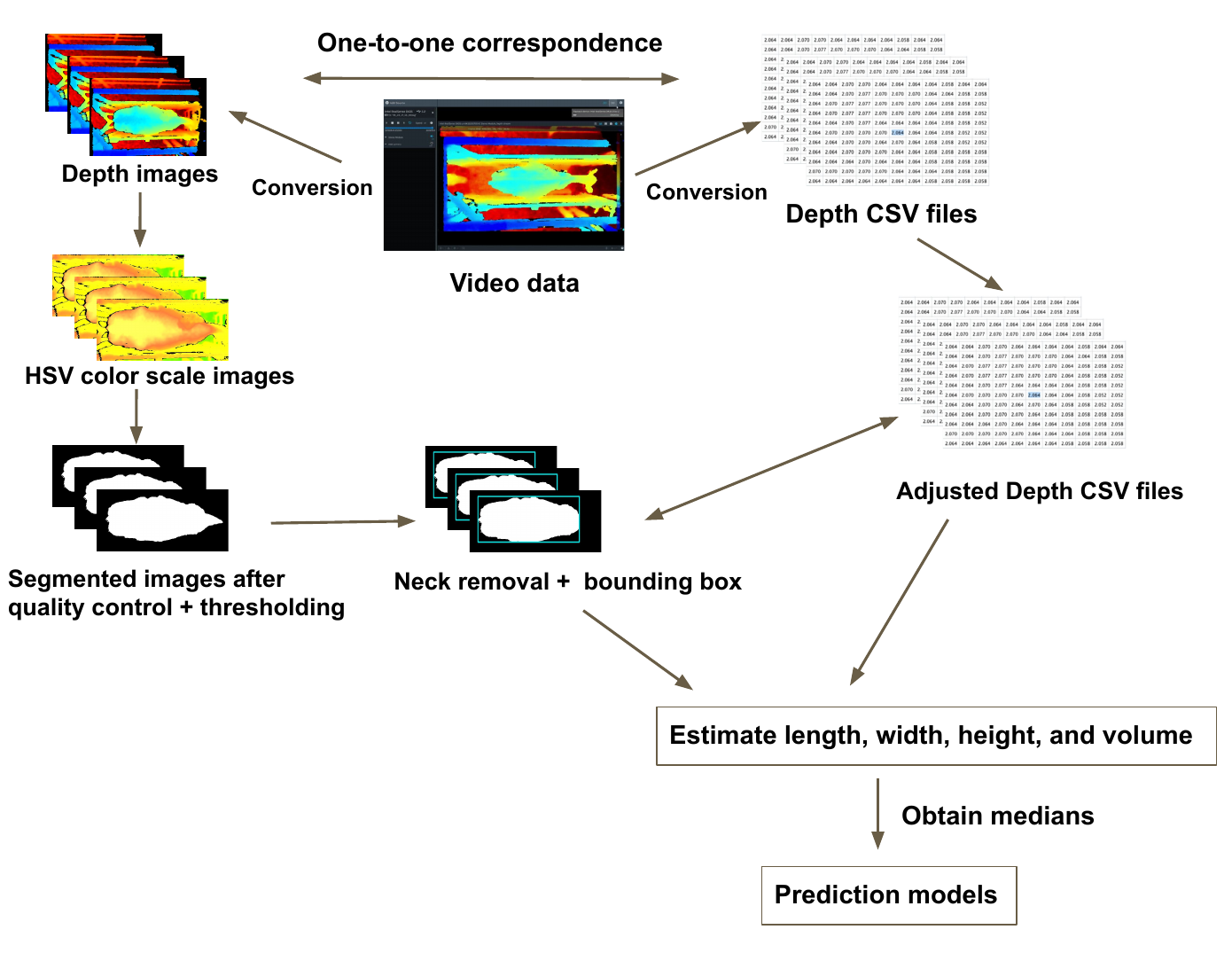}
    \caption{Overview of the biometric approach using thresholding.}
\label{overviewThresholding}
\end{figure}

\begin{figure}[H]
    %\hspace*{-4cm} 
    \centering
    \includegraphics[scale=0.7]{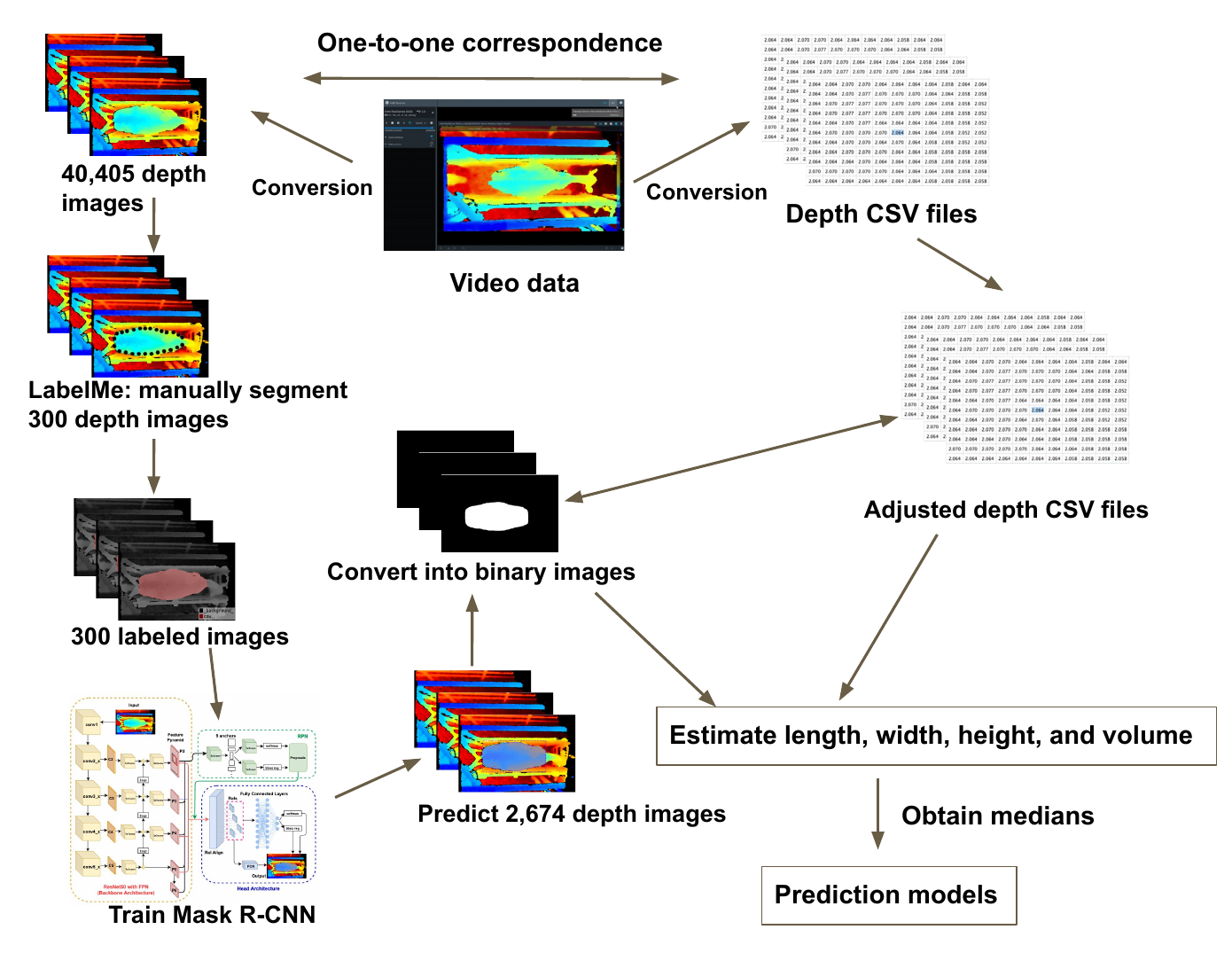}
    \caption{Overview of the biometric approach using Mask R-CNN.}
\label{overviewMaskRCNN}
\end{figure}

Instead of selecting the single best image from each video, all images were processed to estimate biometric features. The biometric features of the cows were estimated from the single-thresholding, adaptive-thresholding, and Mask R-CNN segmented contours. Overviews of the biometric approaches using thresholding and Mask R-CNN are shown in Figures \ref{overviewThresholding} and \ref{overviewMaskRCNN}, respectively. The following four biometric features were extracted from each image.

\noindent
\textbf{Dorsal length} and \textbf{abdominal width}: The dorsal length and abdominal width of the cow in each image were estimated by drawing the bounding box with minimum area using the OpenCV minAreaRect function. 

\noindent
\textbf{Height}: Two methods were evaluated to measure cow height: centroid-based and average-based, to measure cow height. In the first method, the centroid of the cow's contour was determined using the OpenCV moments function. The height of the cow was defined as 2.95 m (height of the camera) minus the distance from the camera to the centroid point using the corresponding CSV depth map file. In the second method, we replaced all zeros (missing data chunks) within contour pixels in the CSV file with the average distance, and the average of all distances between the camera and elements within the entire contour was obtained from the corresponding CSV file. The average cow height was defined as 2.95 m minus the average distance from the camera to the cow contour. 

\begin{figure}[H]
    %\hspace*{-4cm} 
    \centering
    \includegraphics[scale=0.70]{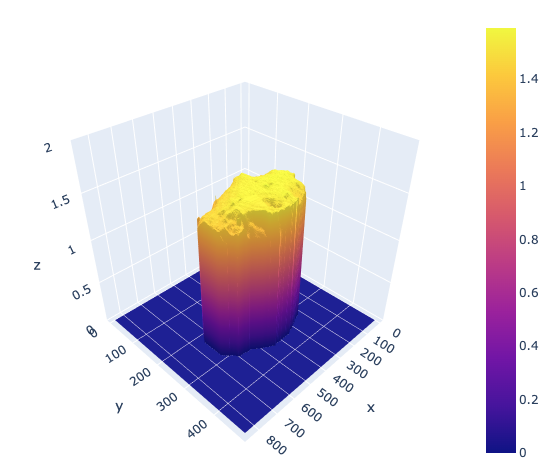}
    \caption{Three-dimensional visualization of the cow volume. The volume of the cow was obtained by integrating the pixels of the cow area over its depth.}
\label{volume}
\end{figure}

\noindent
\textbf{Volume}: Because a 3D depth image captures the curvature of the back, the volume of the cow was obtained by summing the height pixels within the cow contour \citep{nir20183d}. In mathematics, this is analogous to obtaining the volume by integrating the pixels of a cow area over its depth. An example of the estimated cow volume is shown in Figure \ref{volume}. 

Once all images were processed and biometric features were estimated, the medians of dorsal length, abdominal width, height, and volume were calculated from all images to derive the single final estimates for each video. The median, rather than the mean, was chosen to make the final biometric estimates robust to outliers. 

\subsection{Body weight regression model}
After obtaining all four biometric parameters (length, width, height, and volume), we used them as predictors to build regression models to predict dairy cow body weights. We used ordinary least squares (OLS), ridge regression (RR), and least absolute shrinkage and selection operator (LASSO) to estimate model parameters. 

\noindent
\textbf{Ordinary least squares}: OLS was constructed using width, length, height, and volume as predictors. 
\begin{align*}
    \mathbf{y} &= \mathbf{1}\boldsymbol{\mu}+ \mathbf{w}+ \mathbf{l} + \mathbf{h} + \mathbf{v} + \boldsymbol{\epsilon},
\end{align*}
where $\mathbf{y}$ is a vector of body weight (kilograms), $\boldsymbol{\mu}$ is the intercept, $\mathbf{w}$ is a vector of width (pixels), $\mathbf{l}$ is a vector of length (pixels), $\mathbf{h}$ is a vector of height (meters), $\mathbf{v}$ is a vector of volume (cubic meters), $\mathbf{\epsilon}$ is a vector of residuals. The residual sum of squares (RSS) to be minimized for ordinary least square model is
\begin{align*}
\text{RSS}_{\text{OLS}} = (\mathbf{y} - \mathbf{X}\boldsymbol{\hat{\beta}})^{'} (\mathbf{y} - \mathbf{X}\boldsymbol{\hat{\beta}})
\end{align*}
where $\mathbf{X}$ is a matrix including the intercept, length, width, height, and volume; and $\boldsymbol{\beta}$ is a vector of regression coefficients for biometric predictors.

\noindent
\textbf{Penalized regression}: Two types of penalized regression, RR \citep{hoerl1970ridge} and LASSO \citep{tibshirani1996regression}, add a penalty function to the OLS objective function. Specifically, an L2 penalty was used for RR to induce shrinkage, while LASSO adds an L1 penalty to have shrinkage and sparsity (variable selection) properties. The RSS for RR and LASSO are
\begin{align*}
    \text{RSS}_{\text{RR}} &= (\mathbf{y} - \mathbf{X}\boldsymbol{\hat{\beta}})^{'} (\mathbf{y} - \mathbf{X}\boldsymbol{\hat{\beta}}) + \lambda \boldsymbol{\hat{\beta}}^{'}\boldsymbol{\hat{\beta}} \\
    \text{RSS}_{\text{LASSO}} &= (\mathbf{y} - \mathbf{X}\boldsymbol{\hat{\beta}})^{'} (\mathbf{y} - \mathbf{X}\boldsymbol{\hat{\beta}}) + \lambda |\boldsymbol{\hat{\beta}}|
\end{align*}
where $\lambda$ is the regularization parameter, which was fine-tuned using five-fold cross-validation. We implemented OLS, RR, and LASSO using the caret package of R \citep{kuhn2015caret}.

\noindent
\textbf{Linear mixed model}: A linear mixed model (LMM) including random intercepts and random slopes was applied to better capture cow variability or cow longitudinal trajectories of body weight records. The OLS was extended by considering individual cow differences as a random effect (random intercepts) and further assuming that the slopes between body weight and time points vary between cows (random slopes). LMM was implemented using the lme4 package of R \citep{lme4}.

\subsection{Model goodness of fit}
Goodness of fit of OLS, RR, LASSO, and LLM was evaluated using the full data. This measure determines how well a model fits the current data (inference). In other words, goodness of fit assesses how well the model can predict body weight records that have already been used to estimate model parameters. Coefficient of determination ($\text{R}^2$) and mean absolute percentage error (MAPE) were used as goodness of fit evaluation metrics.

\subsection{Cross-validation strategies}
\begin{figure}[H]
    % \hspace*{-3cm} 
    \centering
    \includegraphics[scale=0.7]{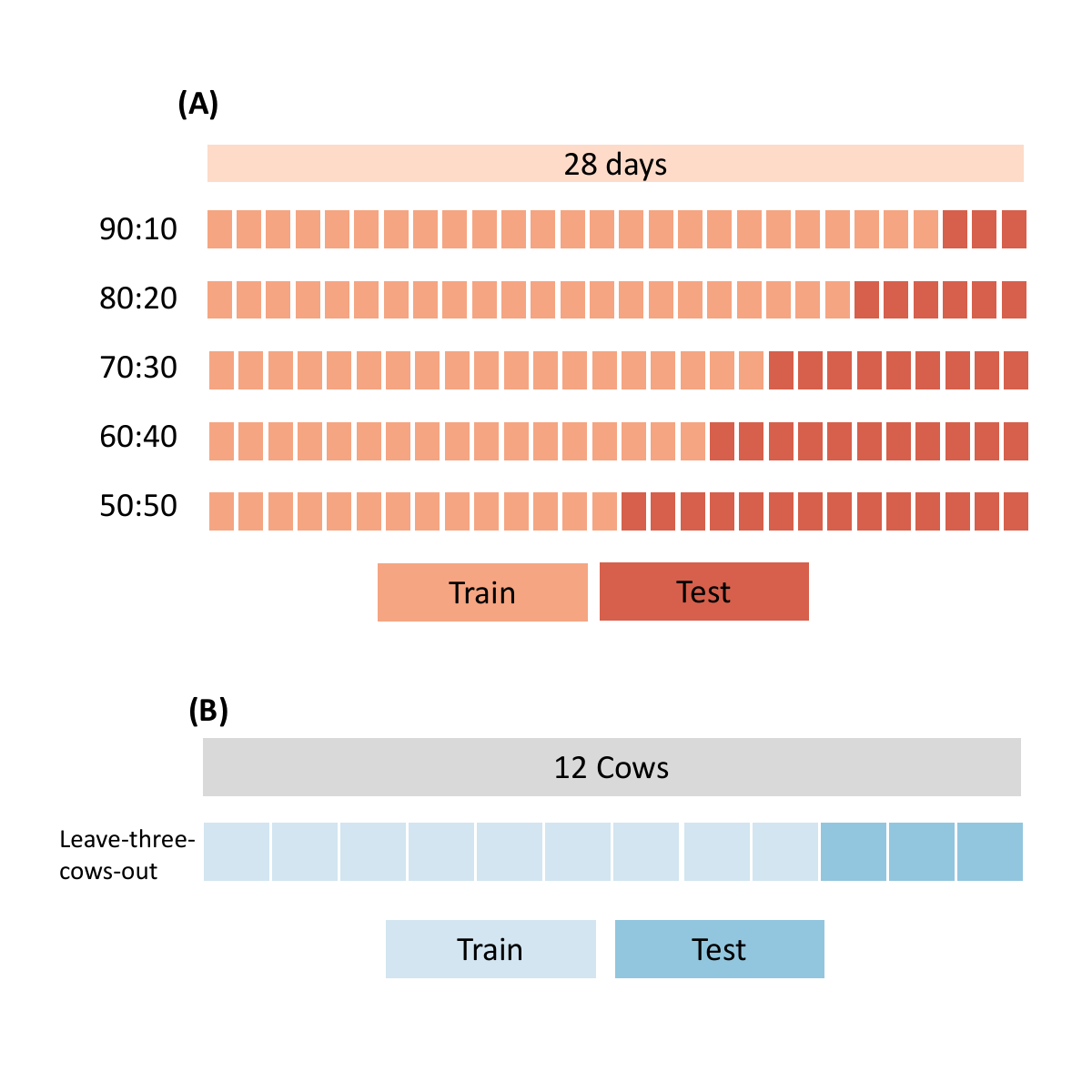}
    \caption{Cross-validation design for body weight prediction. Time series forecasting (A). 90:10: the first 90\% of time points were used to predict the remaining 10\%. 80:20: the first 80\% of time points were used to predict the remaining 20\%. 70:30: the first 70\% of time points were used to predict the remaining 30\%. 60:40: the first 60\% of time points were used to predict the remaining 40\%. 50:50: the first 50\% of time points were used to predict the remaining 50\%. leave-three-cows-out (B). Of 12 cows, 9 and 3 cows were used as the training and testing sets, respectively.}
\label{CVdesign}
\end{figure}

Cross-validation evaluates how well a model can predict new body weight records that were not used to estimate model parameters. Two cross-validation design strategies were used to test model prediction performance: time series forecasting and leave-three-cows-out (Figure \ref{CVdesign}). In time series forecasting, data from earlier time points were trained to predict the body weight of later time points. All data were divided into training and testing sets according to time points. We varied the ratio of training and testing sets to have five scenarios: 90\%:10\%, 80\%:20\%, 70\%:30\%, 60\%:40\%, and 50\%:50\%. Here, 50\%:50\% indicates that the first half of the time points were used to predict the remaining half. In leave-three-cows-out, nine and three cows were used as the training and testing sets, respectively. This corresponds to partitioning the total data into 75\% and 25\%. All possible combinations of training and testing set partitioning were considered $\left(\binom{12}{3} = 220 \right)$. Prediction $\text{R}^2$ and MAPE were used to evaluate the model.

\subsection{Code availability}
The Python and R code used to perform the video processing and regression analyses are available on GitHub (\href{https://github.com/yebigithub/BW_dairy}{https://github.com/yebigithub/BW\_dairy}).

\newpage 
\section{Results}
\begin{figure}[H]
    %\hspace*{-4cm} 
    \centering
    \includegraphics[scale=0.60]{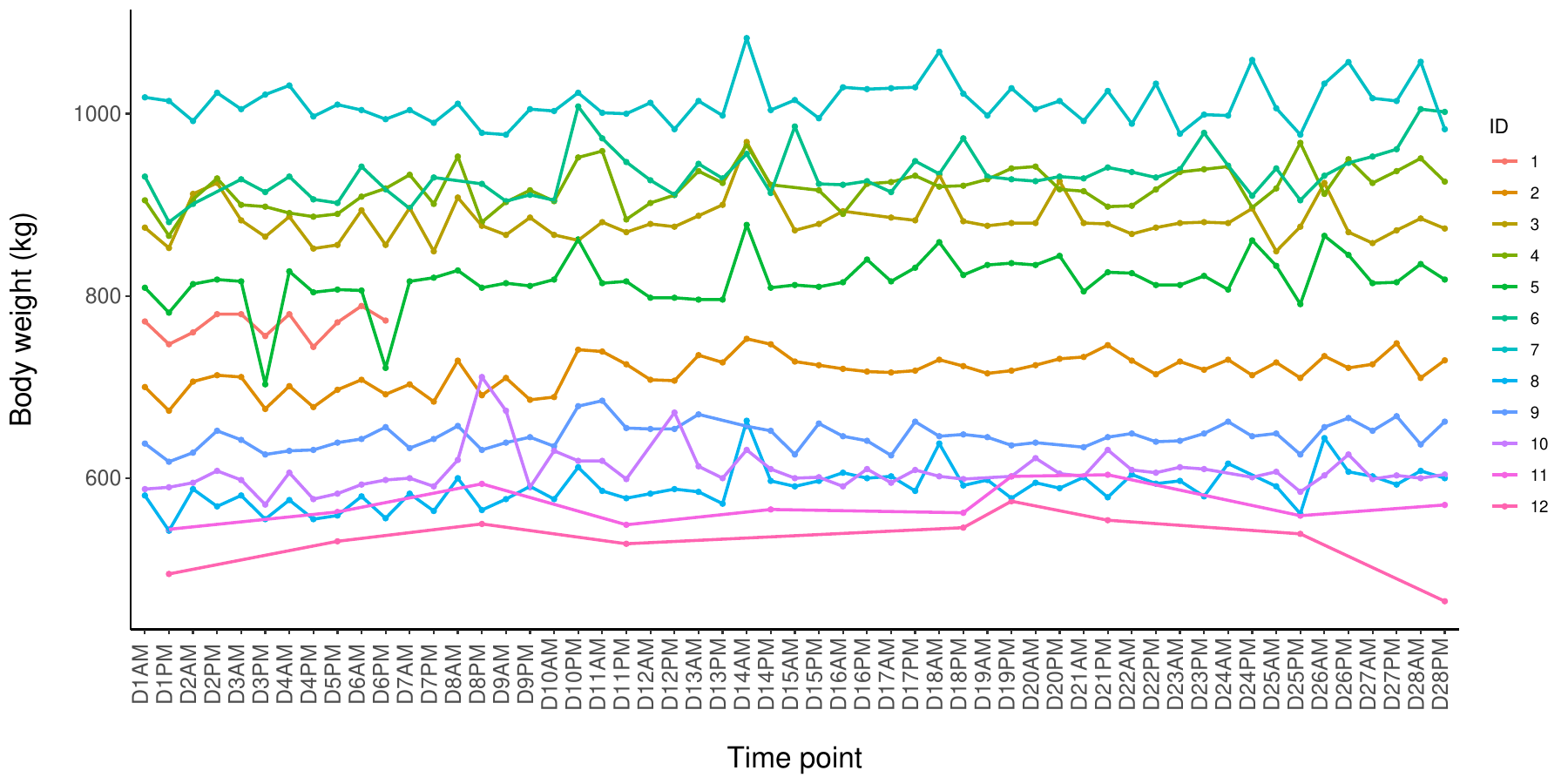}
    \caption{Scale-based cow body weight records across time points. ID 1-10: lactating Holstein cows. ID 11-12: non-lactating Jersey cows.}
\label{BWscatterplot}
\end{figure}

Scale-based body weight measurements automatically collected via the walking-through weighting system across time are shown in Figure \ref{BWscatterplot}. The record taken on Day 5 am was removed because of a systematic error due to the scale. The bottom two cows were non-lactating Jerseys, which had a lower body weight than the Holsteins.

\subsection{Pearson correlation analysis}
\begin{figure}[H]
    %\hspace*{-4cm} 
    \centering
    \includegraphics[scale=0.60]{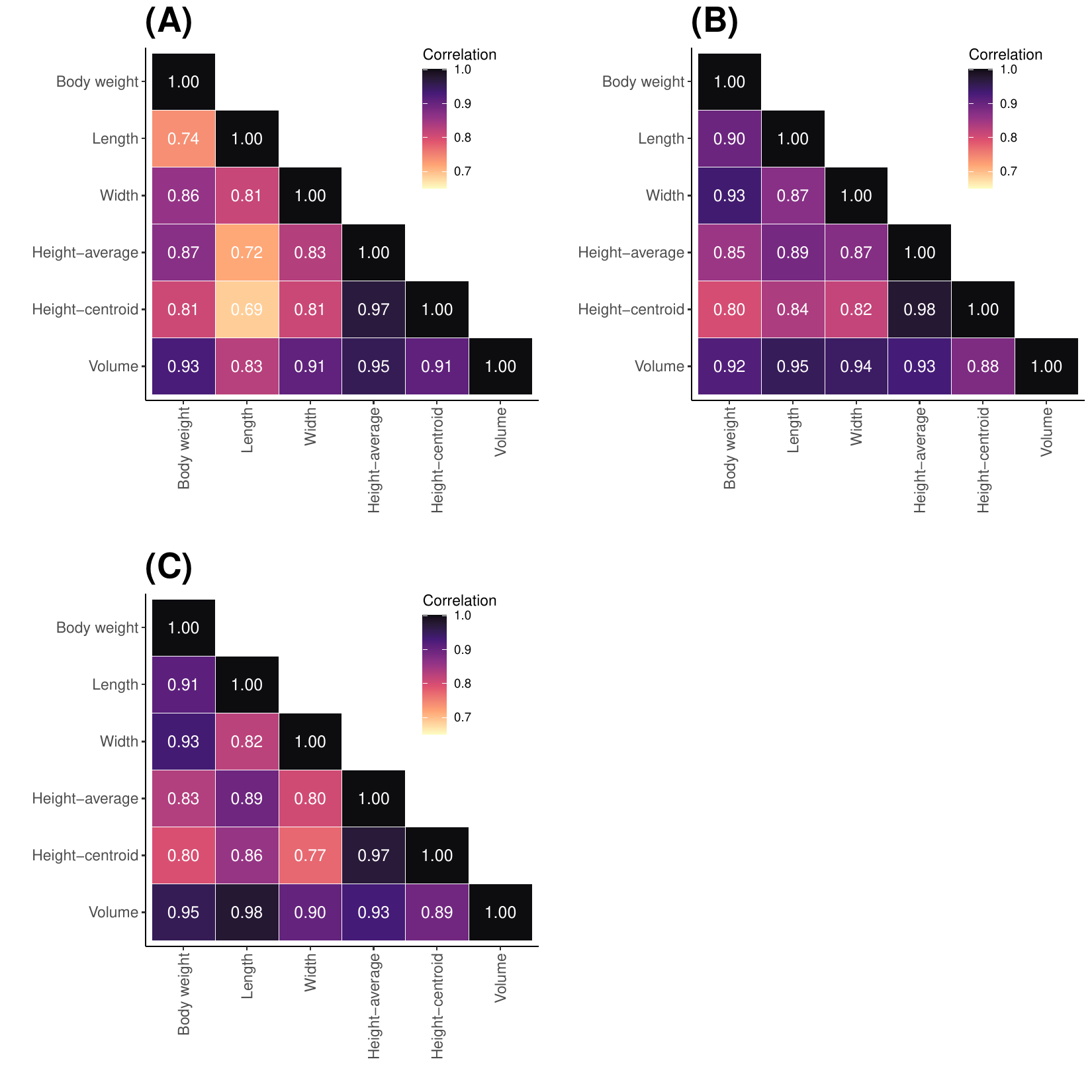}
    \caption{Pearson correlation heat map between scaled-based body weight, length, width, centroid height, average height, and volume averaged across time points. A) single-thresholding, B) adaptive-thresholding, and C) Mask R-CNN.}
\label{Cor_all}
\end{figure}

The confidence score of the trained Mask-RCNN model using 300 images was 0.99, indicating that the model training process was successful. Pearson correlation coefficients between scale-based body weight and five biometric features obtained from the image analysis averaged across time points ranged from 0.74-0.93, 0.80-0.93, and 0.80-0.95 for single-thresholding, adaptive-thresholding, and Mask R-CNN, respectively (Figure \ref{Cor_all}). Volume was the most highly correlated with body weight ($\ge$ 0.93) for all single-thresholding and Mask R-CNN and the second most correlated for adaptive thresholding.  Width was the most correlated biometric feature for adaptive-thresholding and the second most correlated for Mask R-CNN. Length was the least correlated for single-thresholding. Average heights were consistently more correlated than centroid heights. Overall, Mask R-CNN, which provides pixel-level segmentation, resulted in the highest correlation with biometric features and body weight. The patterns of Pearson correlations between scale-based body weight and image-based biometric features were stable over time (Figure S1). Volume and width showed high correlations in all the methods. Again length from single-thresholding was not strongly correlated with body weight. We did not observe major differences in correlations when comparing midnight and noon body weights (Figure S2). 

\subsection{Model goodness of fit}
Based on the correlation analysis, average height rather than centroid height was selected to build regression models for goodness of fit analysis and cross-validation. The extent to which the inferred image-based biometric features explained the observed body weight was evaluated by goodness of fit using the full data. Regardless of the regression models used, the Mask R-CNN segmentation yielded the best goodness of fit according to $\text{R}^2$ and MAPE (Table \ref{GoodnesFit}). Adaptive-thresholding consistently outperformed single-thresholding when OLS, RR, and LASSO were used as regression models. In contrast, the three segmentation methods performed equally well when LMM was fitted ($\text{R}^2$ = 0.98 and MAPE = 1.80\%), which was the best of all. Regarding the regression models, LMM was always the best, followed by OLS, LASSO, and RR.

\subsection{Cross-validation}
Forecasting cross-validation results are shown in Table \ref{CV1Table}. Three regression models, OLS, RR, and LASSO, included only the image-based biometric features as predictors, while LMM included individual cow variability in the model. When the biometric features alone were used to predict body weight, the larger the training data, $\text{R}^2$ and MAPE were slightly larger and smaller, respectively, suggesting that having more past records is useful for predicting future body weight. However, this trend was not observed when single-thresholding biometric features were fitted with RR or LASSO. Mask R-CNN-based segmentation consistently produced the best prediction results. The next best segmentation method was adaptive-thresholding followed by single-thresholding. When $\text{R}^2$ was used to evaluate the prediction performance, OLS and LASSO performed equally well. The prediction performance of RR was slightly lower than OLS and LASSO. The highest $\text{R}^2$ of 0.96 was achieved by the combination of Mask R-CNN and OLS. According to MAPE, OLS performed best, followed by LASSO and RR. Similar to $\text{R}^2$, the lowest MAPE of 3.41\% was obtained by the combination of Mask R-CNN and OLS. When individual cow variability was added to the prediction model using LMM, the prediction performance increased. Similar to the models including biometric features alone (OLS, RR, and LASSO), lower MAPE was observed when historical records were included in the training data. However, there was no obvious difference between the three segmentation methods used. Overall, LMM achieved the highest prediction $\text{R}^2$ of 0.98 and the lowest MAPE of 2.03. 

\begin{figure}[H]
    %\hspace*{-4cm} 
    \centering

    \includegraphics[scale=0.32]{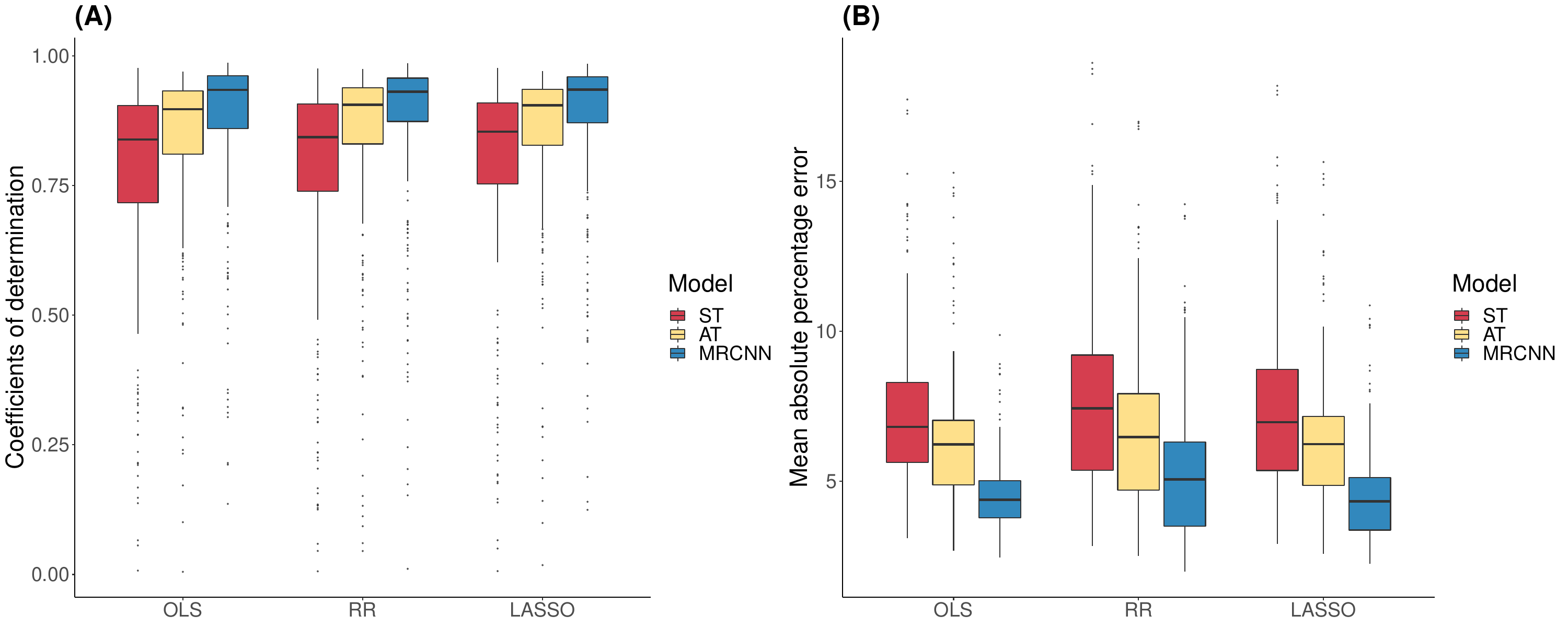}
    \caption{Leave-three-cows-out cross-validation results. A) Prediction coefficient of determination. B) Mean absolute percentage error. ST: single-thresholding. AT: adaptive-thresholding. MRCNN: Mask R-CNN. OLS: Ordinary least squares. RR: Ridge regression. LASSO: Least absolute shrinkage and selection operator.}
\label{CV2Figure}
\end{figure}

The results of the leave-three-cows-out cross-validation are shown in Figure \ref{CV2Figure}. Within each regression model, segmentation-derived parameters from Mask R-CNN consistently gave the best prediction according to $\text{R}^2$ and MAPE. Adaptive thresholding was the second best, followed by single-thresholding.  Although there was not much difference between OLS, RR, and LASSO when evaluated by $\text{R}^2$, OLS produced the best predictions based on MAPE. The combination of Mask R-CNN and OLS produced the highest average $\text{R}^2$ of 0.87 and the lowest average MAPE of 4.53.

\newpage 
\section{Discussion}
In this study, we evaluated three segmentation methods for their preciseness in segmenting body images of dairy cows from background. This segmentation process is the important first step for the application of computer vision in precision livestock farming, because the quality of the biometric features (length, width, height, and volumne) extracted after the image segmentation step greatly influences the subsequent data analysis. We fed the biometric features obtained from the three segmentation methods into four different regression models to evaluate the predictability of longitudinal body weight. The main contribution of the current work was to evaluate the utility of Mask R-CNN instance segmentation and contour extraction for cow body weight prediction in great detail and to compare it with two types of thresholding approaches. 

Of the four biometric measurements, volume was found to correlate most closely with body weight. Although volume is not always used to predict body weight in dairy cows, the importance of volume is consistent with the recent literature \citep{le2019volume,xavier2022use,le2022three}. For instance, \citet{le2019volume} developed a 3D reconstruction system using cloud points to perform body weight prediction on 64 lactating Holstein cows. They found that body weight and volume were most strongly correlated. Theoretically, volume simultaneously accounts for all other biometric features (i.e., length, width, and height). Width was the most correlated and second most correlated with body weight in adaptive-thresholding and Mask R-CNN, respectively. The importance of width is consistent with \citet{song2018automated} reporting among rump length, hip height, and hip width, hip width had the largest correlation with body weight. Height estimates obtained from averages were more correlated with body weight than height estimates obtained from centroids of body regions. This is because, unlike pigs, the height of a cow is often not flat with a curved surface. For this reason, average-based height estimates were used in the regression analyses in our study. 

The quality of biometric features was evaluated from two aspects: goodness of fit and predictability. In the current study, the combination of Mask R-CNN and LMM provided the best model fitness of $\text{R}^2$ of 0.95 and MAPE of 3.57\% when cow variability was not considered. When cow variability was included in the prediction model, $\text{R}^2$ and MAPE improved to 0.98 and 1.80\%, respectively. These values obtained from without or with the inclusion of cow variability were similar or better than previous studies that reported $\text{R}^2$ of 0.80 using six different types of body length and width measurements from eight cows \citep{kuzuhara2015preliminarily}, MAPE of 6.1\% using the volume of 185 cows \citep{hansen2018automated}, $\text{R}^2$ of 0.96 using the rump width, thorax width, and dorsal area of 28 Holstein cows and 27 Holstein heifers \citep{martins2020estimating}, $\text{R}^2$ of 0.92 using the volume of 16 cows \citep{xavier2022use}, $\text{R}^2$ of 0.98 using the volume of 16 cows coupled with LMM including cow variability \citep{xavier2022use}, $\text{R}^2$ of 0.94 using the volume, partial surface area, hip width, chest depth, diagonal length, and heart girth, and body condition score of 16 cows \citep{xavier2022use}, $\text{R}^2$ of 0.98 using the volume, partial surface area, hip width, chest depth, diagonal length, and heart girth and body condition score of 16 cows coupled with cow variability \citep{xavier2022use}, and $\text{R}^2$ of 0.98 using the volume and hip width of 5 heifers \citep{le2022three}. Although the degree of goodness of fit varied among the three segmentation methods, the choice of segmentation method was no longer relevant when LMM was used as the regression model. This suggests that the inclusion of cow variability can compensate for the disadvantages of thresholding compared to Mask R-CNN.  

The first type of prediction scenario was evaluated using forecasting cross-validation. Overall, better prediction results were obtained when a larger number of past records were included in the training sets, suggesting the importance of training data size in forecasting. The best $\text{R}^2$ of 0.96 and MAPE of 3.41\% was achieved by the combination of Mask R-CNN and OLS when cow variability was not included. When cow variability was added to a prediction model, $\text{R}^2$ and MAPE improved to 0.98 and 2.03\%, respectively. Similar to the goodness of fit results, adding cow variability increased predictions and obscured the differences between the segmentation methods. We found that the performance of OLS was better than RR or LASSO. This is probably because shrinkage or variable selection was not strongly needed since the total number of predictors used was only four. Our results show that it is possible to predict future body weight of cows from their past biometric features. One potential application is when ground truth body weight records can only be sparsely collected manually on different days for each cow due to the lack of a walk-over weighing system. In such a case, forecasting can be used to predict body weight on specific days from image data alone.

\begin{figure}[H]
    %\hspace*{-4cm} 
    \centering
    \includegraphics[scale=0.38]{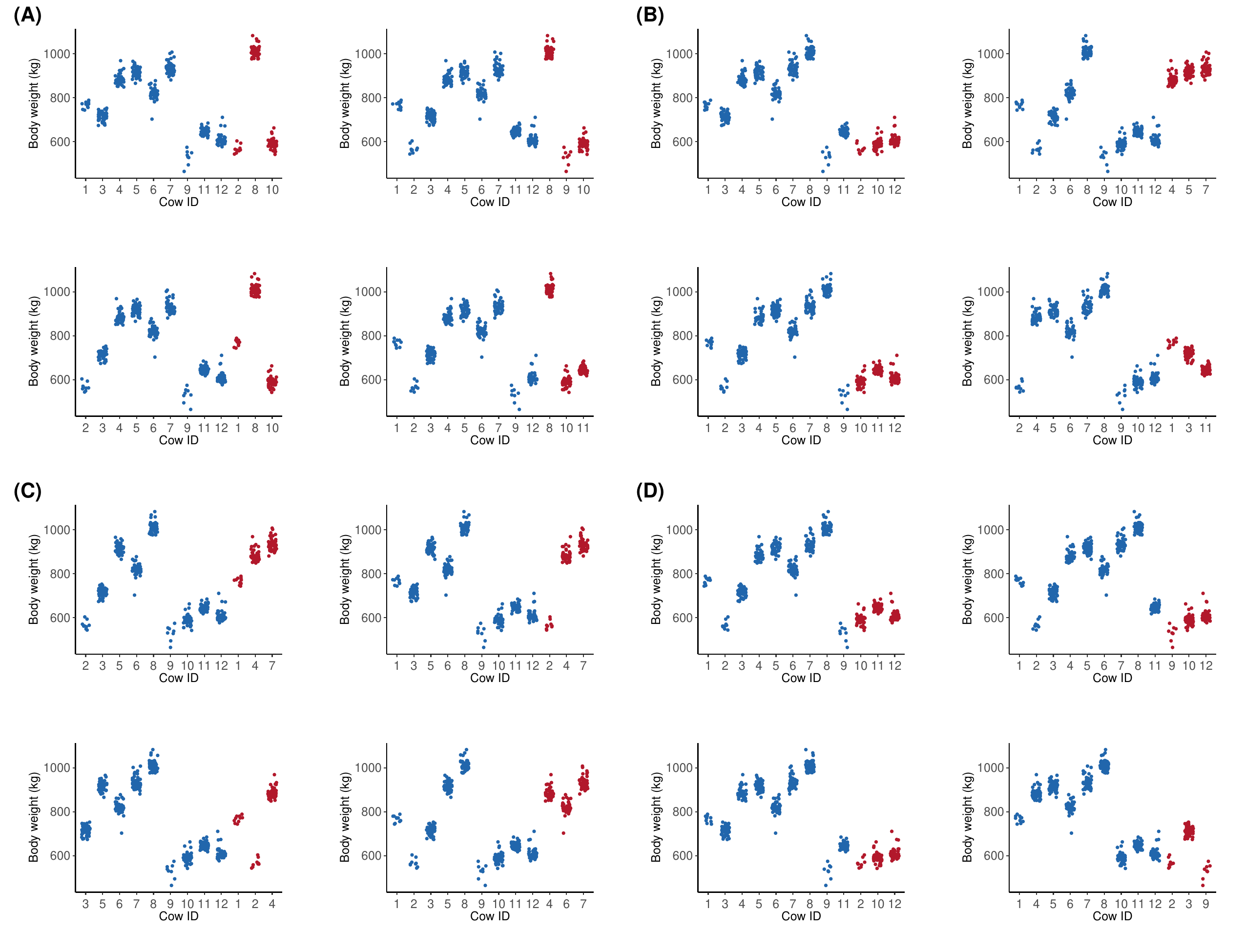}
    \caption{Range of body weight records observed in training and testing set partitioning for leave-three-cows-out cross-validation using Mask R-CNN and ordinary least squares. A) top 4 predictions based on the coefficient of determination. B) bottom 4 predictions based on coefficient of determination. C) top 4 predictions based on mean absolute percentage error. (D) bottom 4 predictions based on mean absolute percentage error.}
\label{Outliers}
\end{figure}

Predictability of new cows not used in the training data was evaluated using leave-three-cows-out cross-validation in the second type of a prediction scenario. Since new cows are predicted in the testing sets, the use of LMM was not feasible in this scenario unless there is genetic information (pedigree or genomics) that allows the cows in the training and testing sets to be connected \citep{baba2020multi}. The combination of Mask R-CNN and OLS achieved the highest average $\text{R}^2$ of 0.86 and the lowest average MAPE of 4.59\%. Overall, our models were able to deliver acceptable predictive performance, but we observed some negative outliers where the predictions were not as successful (Figure \ref{CV2Figure}). These outliers (low $\text{R}^2$ and high MAPE) were further investigated by taking a closer look at the specific partitioning of the training and testing sets that delivered the top 4 predictions and the bottom 4 predictions (outliers) based on $\text{R}^2$ and MAPE (Figure \ref{Outliers}). We found that the bottom 4 predictions always occurred when extrapolation was performed. In such a case, all three cows in the testing sets were in either the upper tail or lower tail of the body weight records. For example, the testing set contained two Jersey cows and one small Holstein cow. In contrast, the top 4 predictions were observed when interpolation was performed where all three cows in the testing sets were within the body weight range of the training set. This contrast occurred because of the relatively small sample size used in this study. It is likely that predictions will be more stable as more data are added to leave-several-cow-out cross-validation. To our knowledge, no previous prediction studies employing cross-validation similar to leave-three-cows-out cross-validation with repeated body weight records were found. Although this prevents a direct comparison of the predictive performance obtained in this study and previous work, our results were within the range or slightly better than previous studies that used larger numbers of cows. For example, $\text{R}^2$ of 0.95 and MAPE of 5.60\% were reported with a hold out method using length, width, area, volume, withers height, hip height, and the additional non-biometric variable age of 107 cows \citep{nir20183d}. Additionally, a MAPE of 7.4\% was obtained in leave-cow-out cross-validation including the hip width of 30 cows \citep{song2018automated} and $\text{R}^2$ of 0.93 and a MAPE of 2.72\% was obtained in repeated random sub-sampling validation replicated 100 times including volume, area, backside width, hip width, heart girth, and wither height of 64 cows \citep{le2019volume}.

In summary, the current study evaluated model goodness of fit and cross-validation predictability of longitudinal body weight records using automatically extracted image-based biometric features. We found that volume followed by width, correlated best with body weight. The combination of Mask R-CNN and OLS or Mask R-CNN and LMM provided the best predictive performance when the model did not include cow variability or when the model included cow variability, respectively. We believe that the availability of the source code used in the current analyses will promote open science in the animal science community.

\newpage 
\section*{Acknowledgments}
This work was funded, in part, by the United States Department of Agriculture National Institute of Food and Agriculture, Hatch project 7000564 and the College of Agriculture and Life Sciences Pratt Endowment at Virginia Tech. We thank Tara Pilonero for making the mounting box to store the depth camera.

\clearpage
\newpage 
\bibliographystyle{apalike} 
\bibliography{CV4CowBW}

\newpage
\section*{Tables}

% Table 1
\begin{table}[hbt!]
\caption{Model goodness of fit.}
\label{GoodnesFit}
\centering
\hspace*{-1cm}
\begin{tabular}{ccccccccc}
\hline
    \multicolumn{1}{c}{Method} &
    \multicolumn{4}{c}{R$^{2^1}$} &  
    \multicolumn{4}{c}{MAPE (\%)$^{2}$}  \\

\cline{1-9}
&OLS$^{3}$ &RR$^{4}$ &LASSO$^{5}$ & LMM$^{6}$  &OLS &RR & LASSO & LMM \\
\cline{2-9}
Single-thresholding
  &0.88  &0.85  &0.87 & 0.98 &5.93   &6.42  &6.04 & 1.80  \\

Adaptive-thresholding 
   &0.91  &0.89  &0.91 & 0.98 &4.72  &5.40 &4.84 & 1.80 \\
 Mask R-CNN
&0.95 &0.94 &0.95 & 0.98  &3.57 &4.36  &3.71 & 1.80 \\

\hline
\end{tabular}
\begin{tablenotes} %(default:normal)
              \item[1]$^{2}$ Coefficients of determination
              \item[2]$^{3}$ Mean absolute percentage error
              \item[3]$^{4}$ Ordinary least squares 
              \item[4]$^{5}$ Ridge regression
              \item[5]$^{6}$ Least absolute shrinkage and selection operator
               \item[6]$^{6}$ Linear mixed model 
 \end{tablenotes}
\end{table}

% Table 2
\newpage
\begin{table}[hbt!]
\caption{Forecasting cross-validation results.}
\label{CV1Table}
\centering
\hspace*{-1cm}
\begin{tabular}{cccccccccc}
\cline{1-10}
    \multicolumn{1}{c}{Method} &
    \multicolumn{1}{c}{Scenario$^{1}$} &
    \multicolumn{4}{c}{R$^{2^2}$} &  
    \multicolumn{4}{c}{MAPE (\%)$^{3}$}  \\

\cline{1-10}
&&OLS$^{4}$ &RR$^{5}$ &LASSO$^{6}$ &LMM$^{7}$ &OLS &RR & LASSO &LMM \\
\cline{3-10}
&90:10 &0.89 &0.86 &0.88 &0.98 &5.94 &6.91 &6.29 &2.10 \\
&80:20 &0.89 &0.86 &0.88 &0.98 &5.82 &6.51 &5.99 &2.17 \\
Single-thresholding
&70:30 &0.87 &0.86 &0.87 &0.98 &6.22 &6.57 &6.29 &2.06 \\
&60:40 &0.88 &0.86 &0.87 &0.98 &6.07 &6.49 &6.13 &2.14 \\
&50:50 &0.88 &0.87 &0.88 &0.98 &6.02 &6.40 &6.03 &3.28 \\
\cline{1-10}
&90:10 &0.93 &0.92 &0.93 &0.98 &4.74 &5.30 &4.75 &2.12 \\
&80:20 &0.91 &0.89 &0.91 &0.98 &5.16 &5.72 &5.22 &2.22 \\
Adaptive-thresholding 
&70:30 &0.91 &0.89 &0.91 &0.98 &5.05 &5.69 &5.14 &2.14 \\
&60:40 &0.91 &0.89 &0.90 &0.98 &5.00 &5.72 &5.16 &2.39 \\
&50:50 &0.91 &0.89 &0.90 &0.98 &4.96 &5.68 &5.10 &3.44 \\
\cline{1-10}
&90:10 &0.96 &0.94 &0.95 &0.98 &3.41 &4.30 &3.53 &2.03 \\
&80:20 &0.96 &0.94 &0.95 &0.98 &3.45 &4.40 &3.66 &2.17 \\
Mask R-CNN
&70:30 &0.96 &0.93 &0.95 &0.98 &3.51 &4.58 &3.87 &2.18 \\
&60:40 &0.96 &0.93 &0.95 &0.98 &3.54 &4.58 &3.84 &2.33 \\
&50:50 &0.96 &0.93 &0.95 &0.98 &3.60 &4.59 &3.90 &3.59 \\
\cline{1-10}

\end{tabular}
\begin{tablenotes} %(default:normal)
              \item[1]$^{1}$ 90:10: 90\% training and 10 \% testing; 80:20: 80\% training and 20 \% testing; 70:30: 70\% training and 30 \% testing; 60:40: 60\% training and 40 \% testing; and 50:50: 50\% training and 50 \% testing
              \item[2]$^{2}$ Coefficients of determination
              \item[3]$^{3}$ Mean absolute percentage error
              \item[4]$^{4}$ Ordinary least squares 
              \item[5]$^{5}$ Ridge regression
              \item[6]$^{6}$ Least absolute shrinkage and selection operator
              \item[7]$^{7}$ Linear mixed model
 \end{tablenotes}
\end{table}

\newpage
\section*{Figures}

\newpage

\newpage

\newpage

\newpage

\newpage

% all days 
\newpage

\newpage

\newpage

\newpage

\newpage

\end{document}

\title{Depth video data-enabled predictions of longitudinal dairy cow body weight using thresholding and Mask R-CNN algorithm}

\author[1]{Ye Bi}
\author[1]{Leticia M.Campos}
\author[2]{Jin Wang}
\author[2]{Haipeng Yu}
\author[1]{Mark D.Hanigan}
\author[1, 3, *]{Gota Morota}
\affil[1]{School of Animal Sciences, Virginia Tech, Blacksburg, VA, 24061 USA}
\affil[2]{Department of Animal Sciences, University of Florida, Gainesville, FL, 32611 USA}
\affil[3]{Center for Advanced Innovation in Agriculture, Virginia Tech, Blacksburg, VA, 24061 USA}

\date{}

\maketitle

\noindent 
$^{*}$ Corresponding author \\ 

\beginsupplement

\newpage
\section*{Figures}
%Figure S1
\begin{figure}[H]
    %\hspace*{-4cm} 
    \centering
    \includegraphics[scale=0.65]{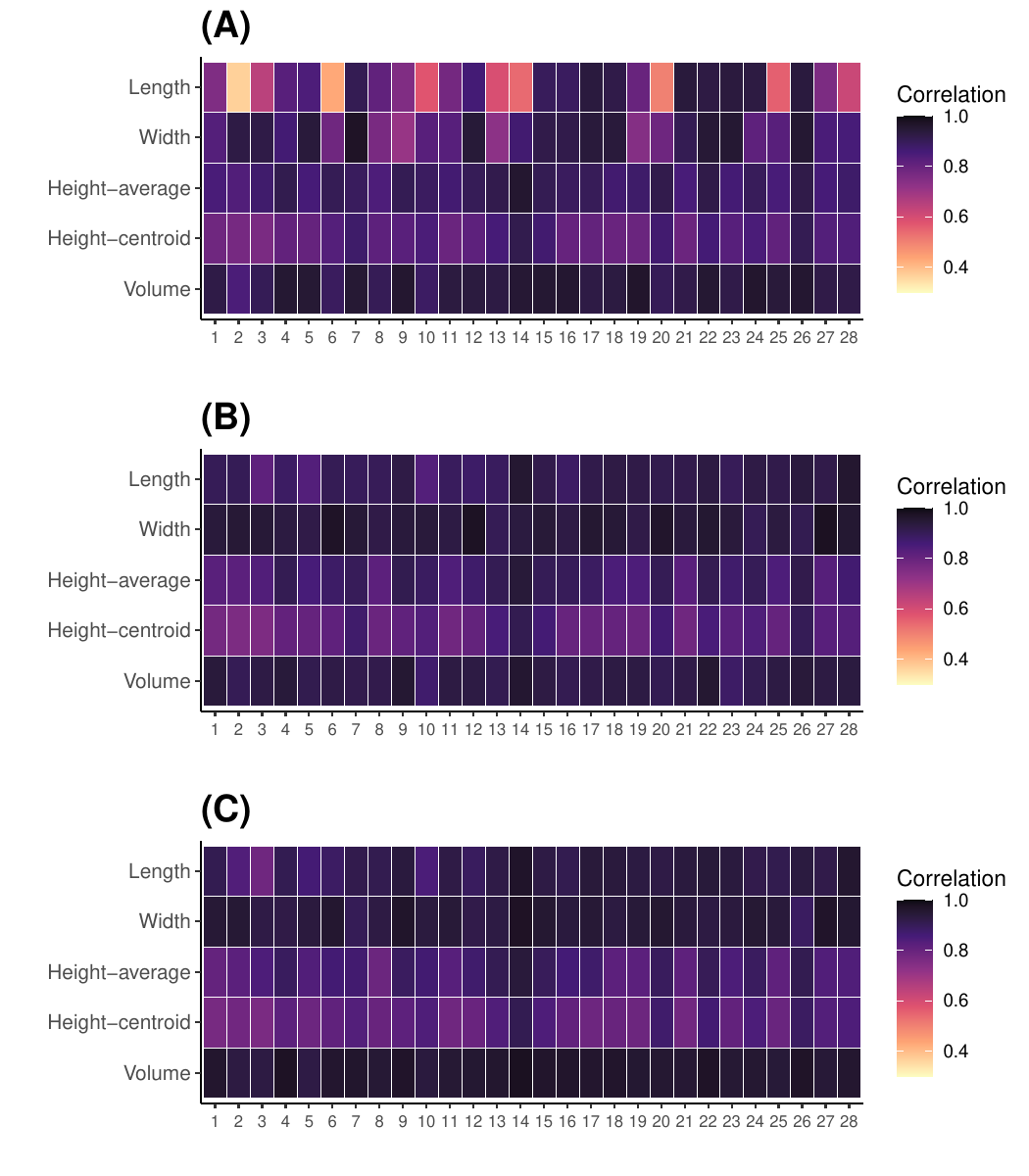}
    \caption{Per day Pearson correlation heat map between scaled-based body weight and biometric features (length, width, centroid height, average height, and volume).  A) single-thresholding, B) adaptive-thresholding, and C) Mask R-CNN.}
\label{Cor_PerDay}
\end{figure}

\newpage
\begin{figure}[H]
    %\hspace*{-4cm} 
    \centering
    \includegraphics[scale=0.5]{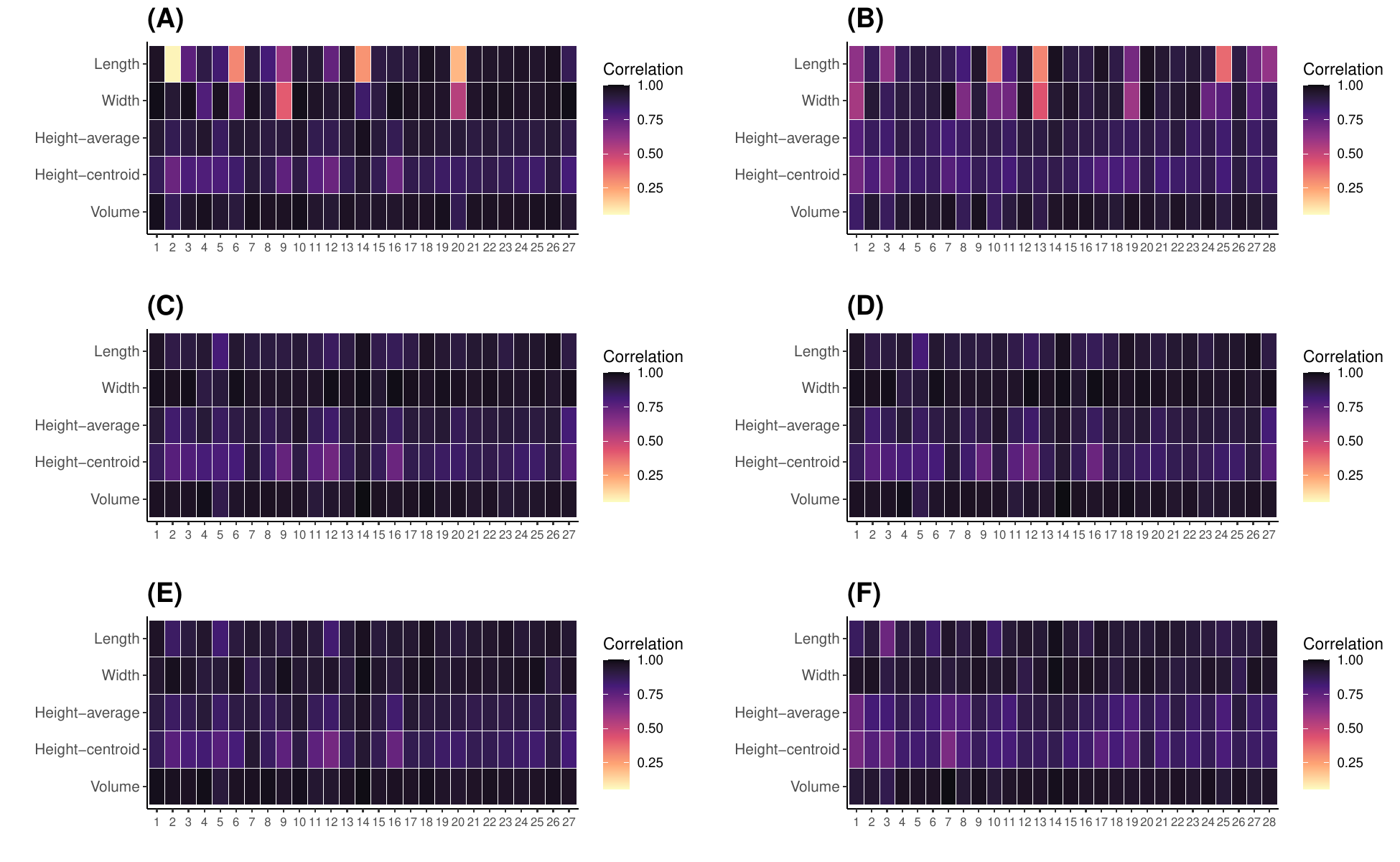}
    \caption{Per day per AM/PM Pearson correlation heat map between scaled-based body weight and biometric features (length, width, centroid height, average height, and volume). A) AM data using single-thresholding, B) PM data using single-thresholding C) AM data using adaptive-thresholding, D) PM data using adaptive-thresholding, E) AM data using Mask R-CNN,  and F) PM data using Mask R-CNN.}
\label{Cor_AMPM}
\end{figure}